\documentclass[review]{elsarticle}

\usepackage{lineno}
\modulolinenumbers[5]

\journal{Artificial Intelligence in Medicine}

\usepackage{graphicx,amssymb}

\usepackage{url,color}
\usepackage{booktabs, array}
\usepackage{multirow}
\usepackage{rotating}
\usepackage{tabularx}
\usepackage{pgfplots}
\pgfplotsset{compat=1.3}

\usepackage[textsize=tiny,colorinlistoftodos]{todonotes}

\newcommand{\chred}[1]{#1}
\newcommand{\chblue}[1]{#1}

\usepackage[normalem]{ulem}

\renewcommand{\sout}[1]{}

\usepackage{algcompatible}
\usepackage{newfloat}
\DeclareFloatingEnvironment[
    fileext=loa,
    listname=List of Algorithms,
    name=ALGORITHM,
    placement=tbhp,
]{algorithm}

\newcommand{\KwIn}[1]{\textit{/*{#1}*/}}
\newcommand{\alg}{\textsc{ExSmi}}

\begin{document}

\begin{frontmatter}

\title{Predicting respiratory motion for real-time tumour tracking in radiotherapy}

\author[TK1,TK2]{Tomas Krilavi\v{c}ius\corref{mycorrespondingauthor}}
\ead{t.krilavicius@bpti.lt}
\cortext[mycorrespondingauthor]{Corresponding author}
\address[TK1]{Baltic Institute of Advanced Technology, Saul\.{e}tekio 15,  LT 10224, Vilnius, Lithuania}
\address[TK2]{Vytautas Magnus University, Vileikos 8, LT 44404, Kaunas, Lithuania}

\author[MI1,TK1]{Indr\.e \v{Z}liobait\.e}
\ead{indre.zliobaite@aalto.fi}
\address[MI1]{Dept. of Computer Science, Aalto University, P.O. Box 15400, FI-00076 Aalto, Finland}

\author[HS]{Henrikas Simonavi\v{c}ius}
\address[HS]{Rubedo Systems, UAB, Jovar\c{u} 2, LT 47193, Kaunas, Lithuania}
\ead{henrikas.simonavicius@rubedo.lt}

\author[LJ]{Laimonas Jaru\v{s}evi\v{c}ius}
\address[LJ]{Hospital of Lithuanian University of Health Sciences Kauno Klinikos, Eiveni\c{u} g. 2, LT 50009, Kaunas, Lithuania}
\ead{laimonas.jarusevicius@kaunoklinikos.lt}

\begin{abstract}
\emph{Purpose}. Radiation therapy is a local treatment aimed at cells in and around a tumor. 
The goal of this study is to develop an algorithmic solution for predicting the position of a target in 3D in real time, 
aiming for the short fixed calibration time for each patient at the beginning of the procedure. 
Accurate predictions of lung tumor motion are expected to improve the precision of radiation treatment by controlling the position of a couch or a beam in order to compensate for respiratory motion during radiation treatment. 

\noindent \emph{Methods}. 
For developing the algorithmic solution, data mining techniques are used. 
A model form from the family of exponential smoothing is assumed, 
and the model parameters are fitted by minimizing the absolute disposition error, and the fluctuations of the prediction signal (jitter). 
The predictive performance is evaluated retrospectively on clinical datasets capturing different behavior (being quiet, talking, laughing), and validated in real-time on a prototype system with respiratory motion imitation.

\noindent \emph{Results}. An algorithmic solution for respiratory motion prediction (called \alg) is designed.
\alg\ achieves good accuracy of prediction (error $4-9$ mm/s) with acceptable jitter values ($5-7$ mm/s), as tested on out-of-sample data.
The datasets, the code for algorithms and the experiments are openly available for research purposes on a dedicated website. 

\noindent \emph{Conclusions}. The developed algorithmic solution performs well to be prototyped and deployed in applications of radiotherapy.
\end{abstract}


\end{frontmatter}


\section{Introduction}
\label{intro}


\label{sec:Introduction}

The goal of radiotherapy treatment is to destroy the tumor and at the same time prevent the healthy surrounding tissues from being damaged \cite{Murphy2004,KeallEtAll2006,SharpJiangShimizu2004,BuzurovicHuangYu2011,ionascu2007internal}. 
Advances in radiotherapy technologies, such as  \emph{intensity modulated or image guided radiotherapy, and stereotactic body radiotherapy}, have made highly conformal and accurate treatment~\cite{BhideNutting2010} possible. 
An important limiting factor to the success of tightly conforming dose distributions is the ability to aim the radiation beam precisely at the target with minimal positional error.    

Therefore, motion management is one of the most active research and development topics in modern radiotherapy, as can be seen from many studies \cite{KeallEtAll2006,BuzurovicYu2012,Abhilash:2013:EMR:2445621.2445740,Abhilash2012,SivaEtAll2013}.

\emph{Intrafraction motion} (motion of the target during treatment) is usually caused by the skeletal muscular, cardiac, gastrointestinal and respiratory systems, the later being responsible for the most of it.

The positions of all the organs in the thorax and abdominal regions are affected by respiration of a patient; however, the organs may move in different ways and various magnitude. 
In addition, the tumor itself may be moving along with the organs, depending on its location and fixation to the surrounding structures. 
The magnitude of the motion highly depends on the location of the tumor and also may vary a lot for individual patients. 
\emph{Lung tumors} can exhibit up to 3 cm motion in the cranio-caudal direction during normal respiration, 
while tumors of other types typically move only a few millimeters or do not move at all~\cite{ErridgeSeppenwoolde2003}.
Movement of lung tumors introduces uncertainty in the positioning. 
To account for this uncertainty the conventional radiation therapy requires larger treatment margins, as it is recommended by the International Commission on Radiation Units and Measurements \cite{ICRU50,ICRU62}. 
Extra margins may lead to large volumes of healthy tissue being destroyed during radiotherapy treatments. 
Therefore, while higher doses of radiation therapy may improve survival rate, \sout{healthy tissue sparing is important due to the concomitant chemotherapy} \chblue{healthy tissue sparing is important to reduce side effects of the organs at risk.}~\cite{KeallEtAll2006}.

To cope with this problem various techniques have been considered \cite{KeallEtAll2006}. 
Active motion compensation \cite{Murphy2004,Ernst13}, 
such as \emph{gated radiotherapy}~\cite{KuboHill1996,ShiratoShimizu1999}, \emph{breath-hold}~\cite{WongSharpe1999} or \emph{tumor tracking}~\cite{AdlerChang1998,RuanFesslerBalter2007,KaletSandisonWu2010,BuzurovicYu2012}
have been introduced into the clinical practice.
However, these techniques have limitations, e.g.\@, the total treatment time significantly increases in case of gated radiotherapy \cite{DietrichTucking2005}, invasive fiducial markers need to be implanted~\cite{KeallTodor2004}, breath-hold works well only in case of compliant patient.
Hence, development of new non-invasive techniques, aimed to controlling respiratory motion in radiotherapy, is an \sout{imperative} \chblue{important} task for the modern radiation oncology. \chblue{Some tracking systems, such as VERO \cite{SolbergEtAll2014-VERO}, that use a beam for positioning and some, like CyberKnife \cite{SeppenwooldeEtAl2007-CyberKnife}  use robotic arm to move linac.}

A generic approach to the compensation for respiratory motion is defined as follows (following ~\cite{Murphy2004}):
(i) determine the current position of the tumor from an external marker position using a computational technique for relating the marker and the tumor \cite{Murphy2004,ionascu2007internal,tsunashima2004correlation,schweikard2000robotic};
(ii) predict the next position of the tumor;
(iii) compensate for the anticipated respiratory motion (e.g. by repositioning the beam); and 
(iv) adapt the dosimetry to the changing configuration of the tumor. 
The current position of a tumor can be determined using external markers \cite{schweikard2000robotic,SharpJiangShimizu2004,tsunashima2004correlation,ionascu2007internal}. 
Once the next position of the tumor is predicted, various techniques can be used to compensate for the respiratory motion \cite{KrilaviciusVitkuteSidlauskas2012,KeallEtAll2006,Murphy2004}, 
e.g.\@ shifting the patient using a robotic-couch, shifting the beam by repositioning the radiation source, redirecting the beam electromagnetically, or changing the aperture of the beam. 

This study focuses on step (ii), i.e.\@, predicting the next position of the tumor from the past observations. Prediction is necessary to overcome delays introduced by tracking system latency.
For predicting the tumor motion a number of predictive modeling techniques have been considered~\cite{Murphy2004,Ernst13,KeallEtAll2006}, such as:
Kal\-man filters~\cite{SharpJiangShimizu2004,Murphy2004,PutraHassMills2008}, 
artificial neural networks (ANN)~\cite{SharpJiangShimizu2004,MurphyPokhrel2009},
state-based probabilistic approaches \cite{KaletSandisonWu2010}, 
local regression\cite{RuanFesslerBalter2007}, 
seasonal autoregressive models (TVSAR) \cite{Ichiji13},
autoregressive moving average models (ARMA)~\cite{RenNishiokaShirato2007}, 
multi-step linear methods (MULIN) \cite{ErnstSchweikard2008}, and
wavelet-based multiscale autoregression (wLMS) \cite{ErnstSchlaeferSchweikard2007}.

While most of the existing studies propose new advanced predictive models, the complete compensation process itself is understudied. 
After selecting an accurate predictive modeling technique, it is far from trivial to put it in operation, for which a full algorithmic solution is required. 
Algorithmic solutions should include step-by-step instructions for automated data pre-processing, model calibration for a given patient, adaptation to potential variation in data arrival rates, confidence estimation and self-diagnosing mechanisms of the model, and potential mo\-de switching (e.g.\@, to a simpler model or no prediction at all). 
The calibration procedure should be done as efficiently as possible in order to minimize preparation time, and maximize utilization of the equipment for treatment. 

This paper proposes a full algorithmic solution for respiratory motion prediction for a selected setup (see sec.~\ref{ssec:DataCollection}), aiming at minimizing the time for model calibration.
The predictive performance is evaluated on clinical datasets off-line and in real-time on prototype system with respiratory motion imitation. 

Several studies develop controllers for motion compensation \cite{Haas12,HerrmanMa2011}, which can be seen as algorithmic solutions, however, 
their focus is on step (iii), i.e. compensating for the anticipated respiratory motion, in Murphy's classification, while our focus is on step (ii), i.e. predicting the next position of the tumor.

The remainder of the paper is organized as follows. 
Sect. \ref{sec:materials} discusses data collection (\ref{ssec:DataCollection}), prediction setting (\ref{ssec:PredictionSetting}), performance criteria (\ref{ssec:PerformanceCriteria}), prediction methods (\ref{sec:predmod}), algorithmic solution (\ref{ssec:Algorithm}) and experimental evaluation (\ref{ssec:ExperimentalEvaluation}). In Section~\ref{sec:results} the performance of the algorithm is evaluated, and in Section \ref{sec:discussion} experimental results are discussed. 
Conclusions and future research directions are presented in Section~\ref{sec:Conclusions}.

\section{Materials and Methods}
\label{sec:materials}

\subsection{Data Collection}
\label{ssec:DataCollection}

\begin{figure}[t]
\centering
\includegraphics[width=\columnwidth]{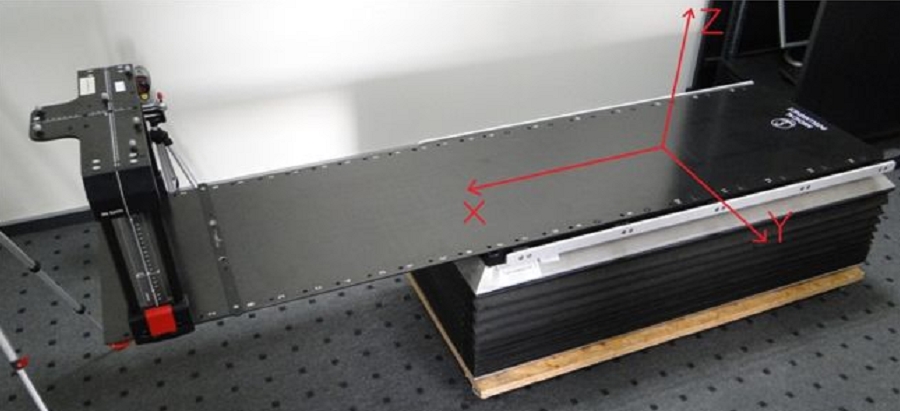}
\caption{General setup for data collection.}
\label{fig:setup-1}
\end{figure}

\begin{figure}[t]
\centering
\includegraphics[width=\columnwidth]{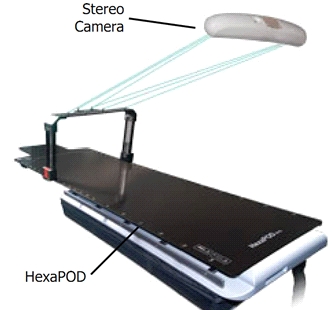}
\caption{Camera position for data collection.}
\label{fig:setup-2}
\end{figure}

Clinical data is collected using an infrared stereo-camera with 60 Hz internal sampling frequency, external markers, HexaPOD evo couch and in-house software. 
Elekta HexaPOD\textsuperscript{\texttrademark} evo RT System\footnote{\url{http://www.elekta.com/healthcare-professionals/products/elekta-oncology/treatment-techniques/positioning-and-immobilization/hexapod-evo-rt-system.html}} (Elekta AB, Stockholm Sweden) 
 is a radiation therapy system setup, with static positioning system iGuide\textregistered 2.0 developed by Rubedo Systems\footnote{\url{http://rubedo.lt/}} (Rubedo Systems, Kaunas, Lithuania). \chred{The system was adapted to collect real-time data by Rubedo Systems.}

The radiation treatment system under consideration consists of several components: \emph{patient setup couch}, in this case the HexaPOD couch\footnote{\url{http://www.elekta.com/assets/Elekta-Oncology/Stereotactic-Radiation-Therapy/case_studies/}} \cite{ChungJinSuh2007}, 
\emph{radiation beam source}, 
usually a \emph{medical linear accelerator} (\emph{linac}), \emph{tracking device}, 
which provides information about the position of the patient and a \emph{controller} that controls the treatment process. Several different control schemes have been  
proposed~\cite{MeyerGuckenbergerWilbert2007,KalokKrilaviciusWan2011,KrilaviciusKalok2011,KrilaviciusVitkuteSidlauskas2012,BuzurovicYu2012}. 

Respiratory motion in HexaPOD is measured by an infrared stereo-camera (NDI Polaris \cite{wiles2004accuracy} (NDI (Northern Digital) International, Ontario, Canada)), that tracks external markers placed on the body of the patient. \chred{We use $1$ mm spatial resolution, and while the camera can provide up-to $0.25$ mm resolution under ideal conditions \cite{wiles2004accuracy}, often it may go up-to $0.6$ mm (with 95\% confidence interval), therefore $1$ mm is the safe choice\footnote{In case of $4-6$ mm it could be insufficient.}.}  
The camera provides position data periodically in frames (and frame numbers).
The timestamps are computed from the sampling frequency of the internal camera, which is $60 Hz$ (a frame in each $16.7$ ms).
Processing delays are negligible ($<1$ ms). 

\chred{The existing setup (provided by Rubedo systems)} is restricted to processing every second (2nd) frame, therefore the effective sampling rate is $33.(3)$, $66.(6)$ or $100$ ($99.(9)$) ms. \chred{See general setup and schematic camera position in Figs. \ref{fig:setup-1} and \ref{fig:setup-2} (this setup is used for the development and testing of iGuide software\footnote{See \url{https://www.youtube.com/watch?v=a4Fqgl6avtA}.}).} 
As a result, the raw incoming data is not completely equally spaced in time, i.e. the time intervals from the second to the sixth or the ninth frame may differ.
We ensure that the data for prediction is equally spaced by resampling the incoming data at a rate that is a multiplier of six frames (which correspond to $100$ ms). 
Due to the same reason, the prediction horizon should also be a multiplier of six frames.
\chred{Prediction horizon isselected based on the specifics of setup, where we have $100$ ms  camera communication delay, and we predict future position $100$ ms ahead to compensate velocity of the couch ($16$ mm/s).}

Ten clinical datasets are used in this study. 
Each dataset includes 3-dimensional observation records with 3 positions per record over time. 
Each dataset records an empty treatment session (no radiation) lasting from $72.617$ to $320.05$ s. 
The datasets have been collected from 3 healthy males aged 20 to 40.
See Table~\ref{table:DataCharacteristics} for further information about the dataset. 
\begin{table*}[h]
\caption{Experimental Data, where SI (superior-inferior) positions are (L-lower, U-upper), body positions (A-abdomen, C-chest and LR (L-left, C-center, R-right), and different states (T-talking, N-normal, O-other (other type of motion), L-laughing); directions: SI (superior-inferior), LR (left-right), AP (anterior-posterior)}
\label{table:DataCharacteristics}
\centering
{\scriptsize
\begin{tabularx}{\textwidth}{XcccX}
\toprule
  \multicolumn{1}{c}{\multirow{2}{*}{\textbf{Signals}}} 
& \multicolumn{1}{c}{\textbf{Max Range}}
& \multicolumn{1}{c}{\textbf{Duration}}
& \multicolumn{1}{c}{\multirow{2}{*}{\textbf{Frames}}}
& \multicolumn{1}{c}{\multirow{2}{*}{\textbf{Experimental setting}}}
\\
& \multicolumn{1}{c}{\textbf{SI, LR, AP (mm)}}
& \multicolumn{1}{c}{\textbf{(s)}}
& 
&
\\ 
\midrule
  201205101519-LACUACUCC-3-T-222	& \centering{19, 4, 23}		& 222.00 & 6658 & 
  lower abdomen center, upper abdomen center, upper chest center, talking\\ 
  201205101522-LACUACUCC-3-N-138 	& \centering{6, 3, 20} 		& 138.00 & 4148 & 
  lower abdomen center, upper abdomen center, upper chest center, normal\\ 
  201205101534-LACUACUCC-3-NO-130 	& \centering{9, 4, 20} 		& 130.00 & 3883 & 
  lower abdomen center, upper abdomen center, upper chest center, normal, other \\ 
  201205101536-LACUACUCC-3-LT-142 	& \centering{29, 14, 31} 	& 142.00 & 4267 &
  lower abdomen center, upper abdomen center, upper chest center, laughing, talking \\ 
  201205101541-LACUACUCC-3-N-130 	& \centering{6, 2, 17} 		& 131.00 & 3919 & 
  lower abdomen center, upper abdomen center, upper chest center, normal\\ 
  201205111055-LACLARUAR-3-N-117 	& \centering{6, 4, 18}		& 117.00 & 3513 & 
  lower abdomen center, lower abdomen right, upper abdomen right, normal \\ 
  201205111057-LACLARUAR-3-O-72		& \centering{40, 10, 45}	& 72.62 & 2178 & 
  lower abdomen center, lower abdomen right, upper abdomen right, other\\ 
  201205181211-LACUACUCC-3-N-320 	& \centering{12, 4, 31}		& 320.05 & 9593 & 
  lower abdomen center, upper abdomen center, upper chest center, normal\\ 
  201205181220-LACUACUCC-3-N-306 	& \centering{20, 5, 36}		& 306.00 & 9176 & 
  lower abdomen center, upper abdomen center, upper chest center, normal \\ 
\bottomrule
\end{tabularx}}
\end{table*}

\subsection{Prediction task}
\label{ssec:PredictionSetting}

Given is a three-dimensional time series  recording the position of an external marker over time. 
The position is given in three coordinates $x$, $y$ and $z$ in millimeters transformed in such a way that $min(x_i)=0, min(y_i)=0$ and $min(z_i)=0$.
Let $r_i =(x_i,y_i,z_i)$ denote the \emph{true position} of a marker at time $i$, and let 
$\hat{r}_i^h = (\hat{x}_i^h,\hat{y}_i^h,\hat{z}_i^h)$ denote the \emph{predicted position} $h$ steps ahead. 
When the horizon $h$ is clear from the context, it will be omitted from the notation.
For brevity we index time series by the index of arrival and not by the time-stamp of arrival. 
The index $i$ refers to the number of the current observation in a sequence from the start of the reading on the current patient.


\subsection{Performance criteria}
\label{ssec:PerformanceCriteria}

\begin{figure}
\centering
\resizebox{\columnwidth}{!}{
\begin{tikzpicture}
\begin{axis}[name=plot1,
height = 3.5 cm, width = 6 cm, axis lines=none,title = {\footnotesize high jitter},ymin = -150, ymax = 150]
\addplot[black, no markers,line width = 2pt] table[x=nn, y=true] {jitter_example.dat};
\addplot[orange, no markers,line width = 1.5pt] table[x=nn, y=pred1] {jitter_example.dat};
\end{axis}
\begin{axis}[name=plot2, at=(plot1.right of south east), anchor=left of south west,
height = 3.5 cm, width = 6 cm, axis lines=none,title = {\footnotesize low jitter},ymin = -150, ymax = 150,
  legend style={legend pos=outer north east, font=\footnotesize, draw=none} ]
\addplot[black, no markers,line width = 2pt] table[x=nn, y=true] {jitter_example.dat};
\addplot[orange, no markers,line width = 2pt] table[x=nn, y=pred2] {jitter_example.dat};
\legend{true signal,prediction}
\end{axis}
\end{tikzpicture}}
\caption{Two predictions giving the same prediction error but different jitters.}
\label{fig:jitter}
\end{figure}
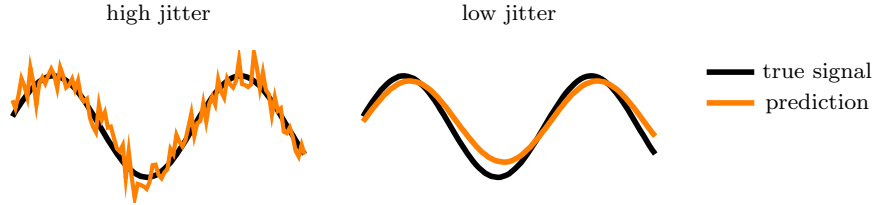

From the operational point of view two performance characteristics are critical: 
predictions should be accurate and the predicted signal should fluctuate as little as possible (have low jitter~\cite{Ernst13}). 
The latter requirement is due to the need for the beamer or the couch to move, following the predicted signal, in order to compensate respiratory motion. 
Following sudden rapid movements of predictions is impractical and may be infeasible \chred{due to mechanical limitations of couch or another tracking device}.
Fig. \ref{fig:jitter} gives two example predictions that lead to the same prediction error, but have different jitters.
A low jitter is preferred from the operational point of view.

Quantitatively the accuracy of predictions can be measured by a straight line distance from the predicted position to the true position in 3-dimensional space (3D). 
For simplicity, distances can be measured in the coordinate units, but for interpretability it is better to transform the coordinates and report results in standard units of length. 
This paper reports prediction errors and jitters in millimeters. 
The prediction error at time $i$ is defined as:
\begin{equation}
e_i = \sqrt{(x_i - \hat{x}_i)^2 + (y_i - \hat{y}_i)^2 + (z_i - \hat{z}_i)^2} = \|r_i-\hat{r}_i\|.
\label{eq:mse}
\end{equation}

The goal is to minimize the error over a treatment session. Since treatment sessions can be of different length, it is practical to look at the mean error over a treatment session: 
\begin{equation}
E = \sum_{i=1}^T e_i / (T\Delta), \end{equation}
where $T$ is the duration of the session in number of frames, and $\Delta$  is the time interval between two frames.

The jumpiness or jitter~\cite{Ernst13} can be measured as the distance the prediction signal travels per time step:
\begin{eqnarray}
j_i &= \sqrt{(\hat{x}_i - \hat{x}_{i-1})^2 + (\hat{y}_i - \hat{y}_{i-1})^2 + (\hat{z}_i - \hat{z}_{i-1})^2} \nonumber\\
 &= \|\hat{r}_i-\hat{r}_{i-1}\|.
\label{eq:jitter}
\end{eqnarray}
For the units (mm) to be interpretable and comparable to the error, in the experiments we will report average jitter and average error per second ($\Delta = 0.1$).

The goal is to minimize jitter over a treatment session. 
Since treatment sessions can be of different length, it is practical to look at the mean jitter over a treatment session: 
\begin{equation}
J = \sum_{i=2}^T j_i/((T-1)\Delta),
\end{equation}
where $T$ is the duration of the session in number of frames, and $\Delta$  is the time interval between two frames.

Note that jitter is minimized when $\hat{r}_i = \hat{r}_{i-1}$ for all $i \in [2,T]$, i.e. the prediction is constant. 
However, in this case no compensation for respiratory movement is possible. 
In practice, a system aims at compensating for respiratory movement, it needs to find a balance between error and jitter.

\subsection{Predictive modeling techniques}
\label{sec:predmod}

We are aiming at developing an algorithmic procedure for real time prediction of respiratory motion. Such an algorithm takes a base model as input, 
and determines when the model should be calibrated, when the actual operation can start, and how to switch between alternative models of different complexity. 

For predicting the tumor motion a range of predictive models have been considered~\cite{Murphy2004,Ernst13,KeallEtAll2006}, as discussed in the introduction.
Our main qualitative criteria for choosing an existing technique for the algorithm  are as follows.
\begin{enumerate}
\item Models need to be fast to calibrate (up to $30-60$ sec) for every next patient, since waiting time is costly.
The number of model design and calibration parameters should be minimal.
\item Models should be able to adapt to changes in respiration rhythm and drifts of the tumor during a session.
\item Models and prediction decisions should be transparent (how predictions are made), so that the technique can be trusted by medical specialists. 
\item Models which are simple to implement on any treatment hardware with minimal usage of external tools are preferable to minimize risks of software errors and dependencies.
\end{enumerate}

Table \ref{tab:survey} provides a summary of considered base models and our assessment against the four qualitative criteria. 
The main limitation of state probabilistic methods (such as Kalman filters or Hidden Markov models) and autoregressive models (such as autoregressive moving average models, regression models fitted using least squares procedure) is that they require relatively large training sample for model calibration before it can be used for predictions, and we are looking for very fast and robust models. More advanced machine learning models (such as neural networks or support vector machines) require even larger training sample sizes, and in addition, the resulting models are so called "black box" models, where it is extremely difficult to trace how the predictions are made.
Therefore, given the focus of our study on fast, interpretable, adaptive and transparent prediction making, we resort to extrapolation and exponential smoothing techniques for our algorithmic solution. The next subsections discuss these two types of techniques in detail. 

\begin{table*}[t]
\caption{Qualitative assessment and selection of base models.}
\label{tab:survey}
\centering
{\scriptsize
\begin{tabularx}{\textwidth}{Xccccc}
\toprule
\multicolumn{1}{c}{\multirow{2}{*}{\textbf{Technique}}} & 
Fast to&
\multirow{2}{*}{\textbf{Adaptive}} &
\multirow{2}{*}{\textbf{Transparent}} &
\textbf{Simple to} & 
\multirow{2}{*}{\textbf{Select}}\\
& 
\textbf{calibrate}&
&
&
\textbf{implement} & 
\\
\midrule
Extrapolation methods		& yes & yes 	& yes 	& yes & \checkmark \\
Exponential smoothing		& yes & yes 	& yes 	& yes & \checkmark \\
State-based probabilistic		& no	& yes/no 	& yes 	& yes & \\
Autoregressive models		& no 	& yes/no 	& yes 	& yes/no & \\
Neural networks			& no 	& yes/no 	& no 	& no & \\
Support vector machines		& no	& no & no & no& \\ 
\bottomrule
\end{tabularx}}
\end{table*}

\subsubsection{Extrapolation methods}
\label{sssec:ExtrapolationMethods}

\paragraph{Extrapolation methods} predict based on the most recent observations. They do not require any calibration and minimum or none parameter settings. 
These methods have very short memory of the past data and hence are inherently adaptive to changes in respiration rhythm or tumor drifts, they are very transparent (easy to explain to a non-specialist) and very simple to implement. 

\paragraph{Persistent prediction} (PP) is the simplest predictor, which predicts that the next signal will be the same as the last observed. No parameters are required. 
\begin{equation}
\hat{x}_{t+h} = x_t,
\end{equation}
here $t$ is the time index and $h$ is the prediction horizon.

Persistent prediction can be considered as a baseline for compensation for respiratory motion. It does not predict pro-actively, but only follows past observations. 

\paragraph{Linear extrapolation} (LE) assumes that the signal will maintain the same velocity and direction as last observed. No parameters are required. 
\begin{equation}
\hat{x}_{t+h} = x_t + (x_t - x_{t-h}).
\end{equation}

\emph{Multi-step linear prediction} (MULIN) \cite{ErnstSchweikard2008} is a generalization over linear extrapolation, 
it takes into account acceleration of the signal of different order. Since the extrapolations may become unstable if the signal is noisy, MULIN uses exponential smoothing moving average of the predictions instead of outputting only the latest prediction.
\begin{eqnarray}
 \hat{x}_{t+h}   		&=& \alpha\left(x_t + \sum_{i=1}^k \delta\left(x_t,h\right)^k\right) + \left(1-\alpha\right)\hat{x}_{t+h-1}\\
 \delta(x_t,h)^1 		&=& x_t - x_{t-h}\\
 \delta(x_t,h)^{i+1} 	&=& \delta(x_t,h)^i - \delta(x_{t-h},h)^i
\end{eqnarray}
where $k \in [1,2,\ldots]$ and $\alpha \in (0,1)$ are user specified parameters. 
In this paper we experiment with the second order MULIN. 
We used the default parameter settings supplied in the implementation made available by the authors\footnote{\url{http://www.rob.uni-luebeck.de/~ernst/dateien/mulin/mulin.m}}.

\subsubsection{Exponential smoothing}
\label{ssec:ExponentialSmoothing}

\emph{Exponential smoothing} is a type of moving average, where the importance of the past observations decreases exponentially. 
Exponential smoothing is not parameter intensive, the only parameter is the speed with which old observations are forgotten. 
Exponential smoothing does not require model calibration for every new patient, it can predict immediately after the start, but a short warm-up period is advisable. 
Just like extrapolation methods, exponential smoothing is inherently adaptive, similarly transparent, and straightforward to implement. 

\paragraph{Simple exponential smoothing} (ES1) makes prediction as the exponentially weighted moving average of the previous observations. 
\begin{equation}
\hat{x}_{t+h} = \alpha x_t+ (1-\alpha)\hat{x}_{t-1},
\end{equation}
for any horizon $h$. 
Here $\alpha \in (0,1)$ is a user defined parameter. If the forgetting factor $\alpha$ is small, then forecasting will have a long memory.
If $\alpha$ is close to one, then forgetting will be fast. 
$\alpha=1$ would mean that we predict the next observation to be the same as the last (PP).
$\alpha = 0$ would give a constant prediction (zero jitter). 
ES1 is equivalent to autoregressive integrated moving average model \cite{Box70} ARIMA(0,1,1).

Simple exponential smoothing does not do well when there is a trend in the data.
 
\paragraph{Double exponential smoothing} (ES2) takes trends into account. 
\begin{eqnarray}
\hat{x}_{t+h}  &=& l_t + hb_t\\
 l_t &=& \alpha x_t + (1-\alpha)(l_{t-1} + b_{t-1})\\
 b_t &=& \beta(l_t - l_{t-1}) + (1-\beta)b_{t-1}
\end{eqnarray}
Here $\alpha \in (0,1)$ and $\beta \in (0,1)$ are user specified parameters. Initialization: $l_0 = x_0$, $b_0=0$.
ES2 is equivalent to ARIMA(0,2,2).

In case of double exponential smoothing for respiratory motion prediction breath cycle will be modeled as short term trends.

The main limitation of this approach is that the prediction will systematically overshoot when the direction of the signal reverses.

\paragraph{Holt-Winters exponential smoothing}, or triple exponential smoothing (ES3), is often used for short term forecasting of seasonal time series \cite{Goodwin10}, as it can handle trends and seasonality. 
Seasonality means that the signal is periodic with a period $p$. We consider ES3 model with additive seasonality component (based on \cite{Hyndman12}). 
\begin{eqnarray}
\hat{x}_{t+h}  &=& l_t + hb_t + s_{t-p+h}\\
 l_t &=& \alpha (x_t - s_{t-p}) + (1-\alpha)(l_{t-1} + b_{t-1})\\
 b_t &=& \beta(l_t - l_{t-1}) + (1-\beta)b_{t-1}\\
 s_t &=& \gamma (x_t - l_{t-1} - b_{t-1}) + (1-\gamma)s_{t-p}
\end{eqnarray}
Here $\alpha \in (0,1)$, $\beta \in (0,1)$ and $\gamma \in (0,1)$ are user specified parameters. Initialization: $l_0 = x_0$, $b_0=0$, $s_0,\ldots,s_{t-p} = 1$.

The original ES3 requires the period to be known and fixed during the model operation. 
The period of a respiratory signal, however, varies even for a single patient, as respiration may become more frequent or slow down over time, the air intake may be delayed due to talking or coughing. We make a stabilizing modification to ES3, 
we use the initial level in estimation of the seasonal component instead of moving average of the level:
\begin{equation}
s_t = \gamma (x_t - l_0) + (1 - \gamma)s_{t-p}.
\end{equation}

We suggest using the parameter values listed in Table~\ref{tab:smoothingpar}, 
which have been found during initial experiments on the training parts of a couple of traces. The testing part of the traces on which the accuracies are reported, was never used for estimating these parameters. 
To minimize the chance of overfitting the training data the parameters are fixed for all the traces. 

We suggest using a fast forgetting for the level (having in mind potential bias of the model and potential drifts), keeping it within a recommended \cite{Hyndman12} restriction $0 < \alpha + \gamma <1$.
The role of the trend component is to estimate long term changes in the average signal level, thus the memory should be long, thus for ES3 $\beta$ should be low.
Since ES2 has no seasonal component, the trend component plays the role of seasonal adjustment, thus $\beta$ needs to be higher than in ES3, but not too high, since in such a case overshooting at turning points may be too large. Since we know that ES2 is biased (data contains seasonality, but we approximate it by the trend component), we need to have a fast forgetting not to propagate model bias therefore $\alpha$ should be high.

\begin{table}
\caption{Recommended parameters for exponential smoothing.}
\label{tab:smoothingpar}
\centering
\begin{tabular}{ccccc}
\toprule
\multirow{2}{*}{\textbf{Model}} &
	 \multicolumn{1}{c}{\textbf{level}} &
	\multicolumn{1}{c}{\textbf{trend}} &
	\multicolumn{1}{c}{\textbf{seasonal}} &
	\multicolumn{1}{c}{\textbf{respiratory rate}}\\
	& \multicolumn{1}{c}{$\alpha$} &
	\multicolumn{1}{c}{$\beta$}  &
	\multicolumn{1}{c}{$\gamma$} & 
        \multicolumn{1}{c}{$p$}\\ 
\midrule
ES1 &    $0.7$    &  &  & \\
ES2 &    $0.7$    &  $0.6$   &    &  \\
ES3 &    $0.7$    &  $0.3$   & $0.3$  & $5.5$ sec \\
\bottomrule
\end{tabular}
\end{table}

%
%

\subsection{Prediction procedure \alg}
\label{ssec:Algorithm}

We propose the following procedure for predicting respiratory motion, called \alg\@, summarized in Algorithm \ref{alg:proposed}.
\alg\ includes online preprocessing outlier removal\footnote{The threshold has been selected based on speed of the coach movement. It is not possible that the couch moves that fast as to produce 1 cm difference between points.} (condition on line 9), online model calibration and switch prediction phase (line 11), a switching mechanism between the main model and a simple, but more robust baseline (line 18), which is based on the most recent performance, taking into account two quantitative criteria - prediction error and jitter (line 18). Linear extrapolation method is considered as a baseline $B$, and exponential smoothing is considered as the main predictive model ($L$).
\begin{algorithm}[h] 
\caption{Predict respiratory signal $h$ steps ahead}
\label{alg:proposed} 
\begin{algorithmic}[1]
{\small
\STATE \textit{incoming observations $r = (x,y,z)$}
\STATE \textit{predictive model form $L$ with design parameters $\theta$}
\STATE \textit{prediction horizon $h$, warm-up $w$ (recommended $w \sim 30$ s)}
\STATE \textit{decay for measuring recent error $d \in (0,1)$ (rec. $d = 0.1$)}
\STATE Initialize model $L_0$ (See Sec.~\ref{sec:predmod} for recommenations)
\STATE Initialize error and jitter counts $E^{L}_0 = 0$, $E^{B}_0 = 0$, $J^{L} = 0$, $J^{B}=0$
\FOR{$t \gets 2,\dots,I$} {\KwIn{from the start to the end of treatment}}
	\STATE receive the latest observation  $r_t$
	\IF{$||r_t - r_{t-1}||<$ 1 cm}
		\STATE update model $L_t = f(L_{t-1},(x,y,z)_t)$
		\IF{$t < w$} {\KwIn{if warmup is over make predictions}}
			\STATE make prediction with $L_t$: $\hat{r}_{t+h}^L$
			\STATE make baseline prediction $\hat{r}_{t+h}^B = r_t + (r_t - r_{t-h})$
			\STATE error $E_t^L = d*error(\hat{r}_t,r_t) + (1-d)E_{t-1}^L$ [Eq. (\ref{eq:mse})]
			\STATE error $E^B_t = d*error(r_{t-h},r_t) + (1-d)E^B_{t-1}$ [Eq. (\ref{eq:mse})]
			\STATE jitter $J_t^L = d*jitter(\hat{r}_t,\hat{r}_{t-1}) + (1-d)J_{t-1}^L$ [Eq. (\ref{eq:jitter})]
			\STATE jitter $J_t^B = d*jitter(r_{t-h},r_{t-h-1}) + (1-d)J^B_{t-1}$ [Eq. (\ref{eq:jitter})]
			\IF {$(E^L_t+J^L_t)>(E^B_t+J^B_t)$} {\KwIn{$L$ performs well}}
				\STATE final prediction by the main model $\hat{r}_{t+h} = \hat{r}_{t+h}^L$
			\ELSE 
				\STATE final prediction by baseline$\hat{r}_{t+h} = \hat{r}_{t+h}^B$
			\ENDIF
		\ENDIF
	\ELSE {\KwIn{$r_t$ is an outlier, ignore}}
		\IF{$t < w$}
			\STATE predict $\hat{r}_{t+h} = \hat{r}_{t+h-1}$
			\STATE set $E_t^L = E_{t-1}^L$, $E^B_t = E^B_{t-1}$, $J_t^L = J_{t-1}^L$, $J^B_t = J^B_{t-1}$
		\ENDIF	
	\ENDIF
	\STATE adjust the beamer \KwIn{out of the scope of this paper}
\ENDFOR
}
\end{algorithmic}
\end{algorithm}

At the time of model switch (line 18) both models are well warmed up, and estimates of the most recent performance are available. 
\chblue{We select the model demonstrating the lowest \emph{recent} prediction error and jitter of the two and apply a fading factor $\alpha$ to the running estimates of the performance to ensure that the most recent performance is accounted with more weight, while considerng not-so-recent performance history with lower weight.} IT helps to minimize the risk of sudden jumps in prediction error or jitter, when the \chblue{predictors are switched}. In the next section we experimentally analyze the performance of the proposed approach. 

A recent study \cite{Ernst11} aims at classifying the patients into predictable and unpredictable, in order to decide whether motion compensation should be used at all, which is conceptually similar to our approach, but there are several key differences. While the authors consider whether motion compensation should be used or not, we do not question the applicability of motion compensation, but dynamically switch between models of different complexity depending on the noisiness of the signal. Moreover, their approach is to decide regarding each patient before commencing the treatment, while our approach is intended for real time, and the outcome during each new treatment session may be different.

\subsection{Protocol for experimental analysis}
\label{ssec:ExperimentalEvaluation}

We experimentally analyze the performance of the base models selected in Section \ref{sec:predmod} and the proposed algorithmic solution in the following settings. 
New observations arrive every $100$ ms and the required prediction horizon is $200$ ms ahead ($h=2$). 
The warm up period is $30$ sec, which is $300$ samples ($w=300$). 
Prediction errors and jitters are reported as averages from observation $301$ until the end of the treatment session. 
We first test the prediction methods stand alone, and then test a selected prediction method inside the proposed algorithm. 

All the experiments are performed using in-house produced 
MATLAB\textregistered\ code, available at \url{http://datasets.bpti.lt/radiotherapy}.

\section{Results}
\label{sec:results}

Our experimental analysis consists of two parts: firstly, we experimentally evaluate the performance of alternative prediction methods in terms of prediction error and jitter, and secondly, we experimentally analyze the performance of the proposed algorithmic solution. 

\subsection{Predictive performance of base models}
\label{ssec:Performance}

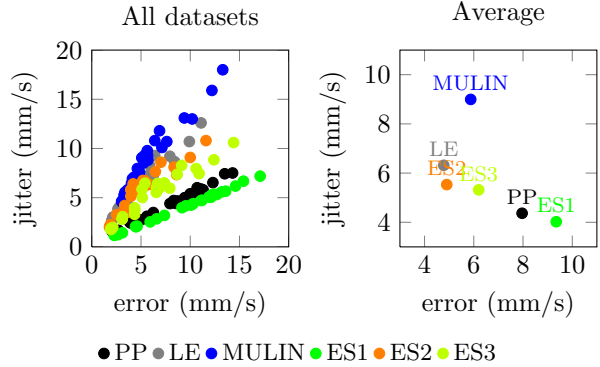
\begin{figure}
\centering
\begin{tikzpicture}
\begin{axis}[name=plot1, 
height = 4.2 cm, width = 4.2 cm, xmin = 0, xmax = 20, ymin = 0, ymax = 20,
title = All datasets, xlabel = error (mm/s), ylabel = jitter (mm/s),
legend style={legend pos=outer north east, font=\small, draw=none},
legend columns=-1, legend entries={PP,LE,MULIN,ES1,ES2,ES3}, legend to name=named5]
\addplot[black, only marks] table[x=eLP, y=jLP] {results1.dat};
\addplot[gray, only marks] table[x=eLE, y=jLE] {results1.dat};
\addplot[blue, only marks] table[x=eMULIN, y=jMULIN] {results1.dat};
\addplot[green, only marks] table[x=eES1, y=jES1] {results1.dat};
\addplot[orange, only marks] table[x=eES2, y=jES2] {results1.dat};
\addplot[lime, only marks] table[x=eES3, y=jES3] {results1.dat};
\legend{PP,LE,MULIN,ES1,ES2,ES3}
\end{axis}
\begin{axis}[name=plot2, at=(plot1.right of south east), anchor=left of south west,
height = 4.2 cm, width = 4.2 cm, xmin = 3, xmax = 11, ymin = 3, ymax = 11,
title = Average, xlabel = error (mm/s), ylabel = jitter (mm/s),nodes near coords]
\addplot [black, mark = *, nodes near coords=\footnotesize{PP}] coordinates {(7.963,   4.367)};
\addplot [gray, mark = *, nodes near coords=\footnotesize{LE}] coordinates {(4.780,   6.317)};
\addplot [blue, mark = *, nodes near coords=\footnotesize{MULIN}] coordinates {(5.875,   8.995)};
\addplot [green, mark = *, nodes near coords=\footnotesize{ES1}] coordinates {(9.350,   4.022)};
\addplot [orange, mark = *, nodes near coords=\footnotesize{ES2}] coordinates {(4.902,   5.532)};
\addplot [lime, mark = *, nodes near coords=\footnotesize{ES3}] coordinates {(6.200,   5.318)};
\end{axis}
\end{tikzpicture}
\ref{named5}
\caption{Predictive performance of alternative base models. (left) Each dot is one time series, (right) average performance.}
\label{fig:results}
\end{figure}

Figure \ref{fig:results} depicts prediction errors and jitters of the base models, selected in Section \ref{sec:predmod}. 
On the left plot each dot is one time series (recall that each dataset includes three positions, that is, three time series). 
We can see from the left plot that the selected models provide a variety of errors and jitters, indicating that some of the signals are more difficult to predict than the others. However, dots of the same color (the same base model) appear in elongated clusters, suggesting that there may be a trade-off between error and jitter achieved by different models, that is, a gain in error increases jitter and the other way around.
Some models may be better at minimizing error, others at keeping the jitter low. 
The right plot presents the average overall time series for each model. 
We see that ES1, PP, ES3 and LE demonstrate nearly a linear trade-off between jitter and error with ES1 showing the lowest jitter and LE showing the lowest error. 
MULIN demonstrates a reasonable error, but the jitter is much worse than that of the other models. 
ES2 achieves nearly the same error as LE, but has a lower jitter, therefore, we select ES2 as the primary base model for our algorithmic solution.
The performance of a constant prediction, which achieves zero jitter, is not plotted since the error ($51$mm/s) is too far off the scale of the plot.

\subsection{Visual analysis}
\label{ssec:VisualAnalysis}

Figure \ref{fig:perfzoom} plots the predictions of the compared methods on a snapshot of the first coordinate from
the experiment 2012\-05181211-UAC-1-N-320-6. We can see that PP and ES1 have a regular delay in predictions, LE and ES2 overshoot at peaks, MU\-LIN and ES3 follow the signal reasonably well, but MULIN is too spiky (high jitter) and ES3 occasionally makes sudden errors. 
Based on this visual analysis ES2 or LE are preferred methods.
\newcommand{\pwidth}{5.5 cm}
\newcommand{\ymn}{-17}
\newcommand{\ymx}{-3.5}
\begin{figure}
\centering

\begin{tikzpicture}
\matrix{
\begin{axis}[y=0.1cm, width = \pwidth, axis lines=none, title = \footnotesize{PP}, ymin=\ymn,ymax=\ymx, xmin=1,xmax=111,
legend style={legend pos=outer north east, font=\small, draw=none},
legend columns=-1, legend entries={\footnotesize{true signal},\footnotesize{prediction}}, legend to name=named6]
\addplot[black, no markers] table[x=nn, y=true] {prediction_example.dat};
\addplot[orange, no markers] table[x=nn, y=eLP] {prediction_example.dat};
\end{axis}
& \begin{axis}[
y=0.1cm, width = \pwidth, axis lines=none, title =  \footnotesize{LE}, ymin=\ymn,ymax=\ymx, xmin=-9,xmax=101]
\addplot[black, no markers] table[x=nn, y=true] {prediction_example.dat};
\addplot[orange, no markers] table[x=nn, y=eLE] {prediction_example.dat};
\end{axis}\\
\begin{axis}[
y=0.1cm, width = \pwidth, axis lines=none,title = \footnotesize{MULIN}, ymin=\ymn,ymax=\ymx, xmin=1,xmax=111]
\addplot[black, no markers] table[x=nn, y=true] {prediction_example.dat};
\addplot[orange, no markers] table[x=nn, y=eMULIN] {prediction_example.dat};
\end{axis}
& \begin{axis}[
y=0.1cm, width = \pwidth, axis lines=none, title = \footnotesize{ES1}, ymin=\ymn,ymax=\ymx, xmin=-9,xmax=101]
\addplot[black, no markers] table[x=nn, y=true] {prediction_example.dat};
\addplot[orange, no markers] table[x=nn, y=eES1] {prediction_example.dat};
\end{axis}\\
\begin{axis}[
y=0.1cm, width = \pwidth, axis lines=none,title = \footnotesize{ES2}, ymin=\ymn,ymax=\ymx, xmin=1,xmax=111]
\addplot[black, no markers] table[x=nn, y=true] {prediction_example.dat};
\addplot[orange, no markers] table[x=nn, y=eES2] {prediction_example.dat};
\end{axis}
& \begin{axis}[
y=0.1cm, width = \pwidth, axis lines=none, title = \footnotesize{ES3}, ymin=\ymn,ymax=\ymx, xmin=-9,xmax=101]
\addplot[black, no markers] table[x=nn, y=true] {prediction_example.dat};
\addplot[orange, no markers] table[x=nn, y=eES3] {prediction_example.dat};
\end{axis}\\
};
\end{tikzpicture}
\ref{named6}
\caption{Performance on a snapshot of experiment 201205181211-UAC-1-N-320-6.}
\label{fig:perfzoom}
\end{figure}
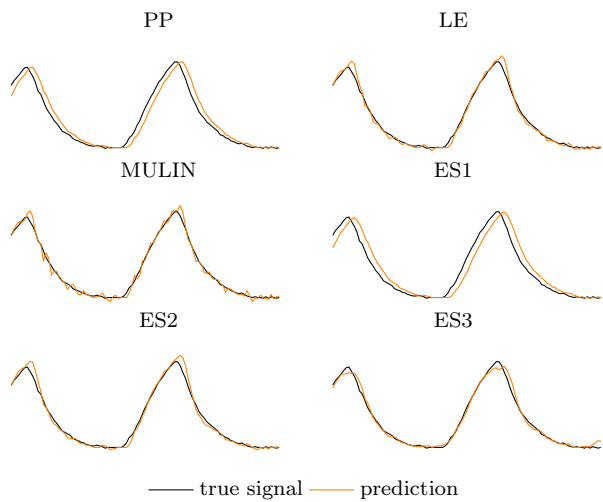
Figure \ref{fig:unfolded} compares the jitters produced by different methods. 
Each line shows how much the beam would need to travel in $10$ seconds if the predictions were followed. 
We see that all methods are comparable in terms of jitter except for MULIN, which produces substantially larger jitter. 
\newcommand{\lw}{1.5pt}
\begin{figure*}
\begin{tikzpicture}
\begin{axis}[ %
	axis y line=none,
	axis x line=bottom,
	xmin=0, xmax=110,
	ymin=-1, ymax=5.4,
	height=4cm,
	x=0.1cm,
	title style={at={(1.12,-0.2)}},
	title={jitter (mm/10s)},
]
\addplot[orange, no markers,line width = \lw] coordinates{(0,4) (53.99,4)};
\node at (axis cs:53.99,4) [orange,anchor= west] {{\footnotesize PP}};
\addplot[orange, no markers,line width = \lw] coordinates{(0,1) (68.92,1)};
\node at (axis cs:68.92,1) [orange,anchor= west] {{\footnotesize LE}};
\addplot[orange, no markers,line width = \lw] coordinates{(0,0) (92.91,0)};
\node at (axis cs:92.92,0) [orange,anchor= west] {{\footnotesize MULIN}};
\addplot[orange, no markers,line width = \lw] coordinates{(0,5) (50.81,5)};
\node at (axis cs:50.81,5) [orange,anchor= west] {{\footnotesize ES1}};
\addplot[orange, no markers,line width = \lw] coordinates{(0,2) (62.84,2)};
\node at (axis cs:62.84,2) [orange,anchor= west] {{\footnotesize ES2}};
\addplot[orange, no markers,line width = \lw] coordinates{(0,3) (60.01,3)};
\node at (axis cs:60.01,3) [orange,anchor= west] {{\footnotesize ES3}};
\end{axis}
\end{tikzpicture}
\caption{\label{fig:unfolded} How much the couch would need to travel in 10 s if the predicted signals are followed (including both jitter, and respiratory motion), averaged over experiment 201205181211-UAC-1-N-320-6.} 
\end{figure*}
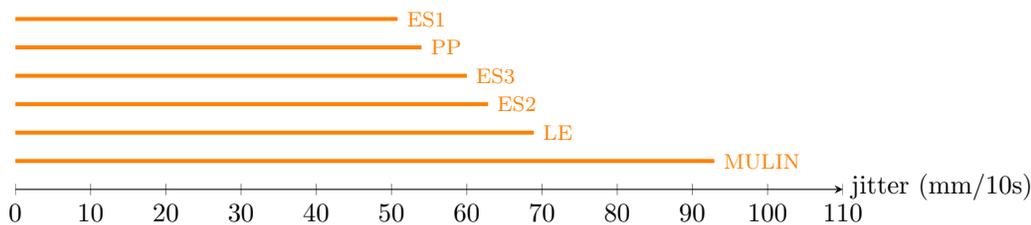

Next we look at the scatter of predictions in space from a patient's perspective.  
Assuming that the bed can perfectly track the predictions, we plot where the beam will hit in 2D with respect to the true target. 
For that we subtract the true signal from the prediction, this way the true target is always $(0,0)$. 
Intuitively, in order not to cause unnecessary harm to the patient,  deviations of predictions from the target $(0,0)$ should be as small as possible and there should be no far outliers. 
Moreover, the errors should be distributed around the target $(0,0)$ as evenly as possible, not concentrated in one or a few spots.

\begin{figure*}
\centering
\includegraphics[width=\textwidth]{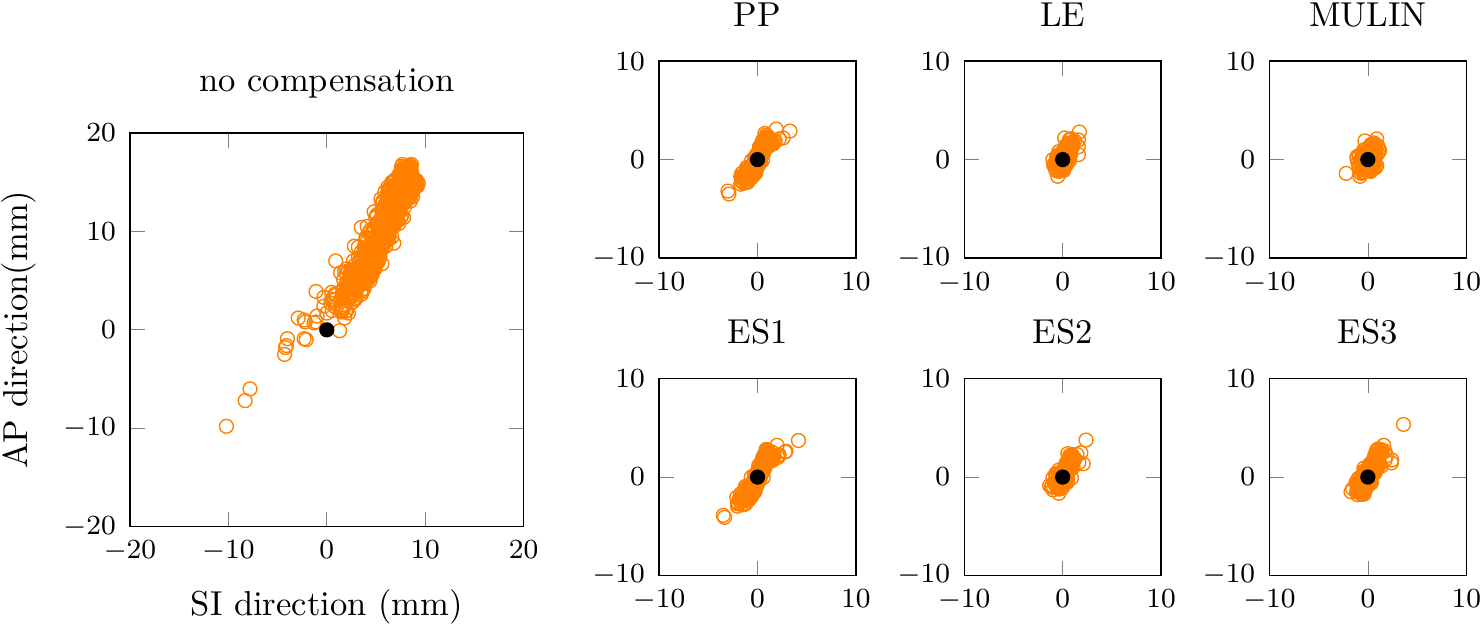}
\caption{Scatters of predictions on the testing range of 201205181211-UAC-1-N-320-6.} 
\label{fig:scatter}
\end{figure*}

Fig. \ref{fig:scatter} plots the scatters of predictions for the same experiment 201205181211-UAC-1-N-320-6 ) \chred{(resampled 201205181211-UAC-1-N-320)} in 2D (x and z coordinates). 
We see that all the six methods produce predictions that are reasonably close to the true target, as compared to no compensation. However, PP and LE1 have the strongest tendency to make concentrated errors, meaning that particular two spots on the upper right and lower left sides from the target may be burned due to prediction latency. \chblue{We would like to notice that in this research tumor is treated as a point (centroid) representing a 3D volume\sout{, therefore errors that are concentrated near the target potentially would be a part of tumor volume}.}
\sout{However, we treat a tumor as a point, while it has some 3D dimensions, therefore error, that is distributed just around the central point maybe not so bad, i.e. it would still may remain in the area of tumor.}
 
\newcommand{\cmm}{20}
\pgfplotsset{x tick label style = {font=\scriptsize},y tick label style = {font=\scriptsize},compat=1.3}

\subsection{Performance of the proposed algorithm}
\label{ssec:PerformanceAlgorithm}

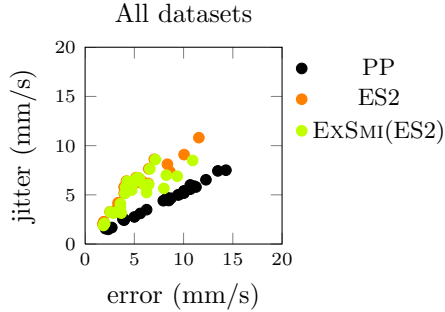
\begin{figure}
\centering
\begin{tikzpicture}
\begin{axis}[height = 4.2 cm, width = 4.2 cm, xmin = 0, xmax = 20, ymin = 0, ymax = 20,
title = All datasets, xlabel = error (mm/s), ylabel = jitter (mm/s),
legend style={legend pos=outer north east, font=\small, draw=none}]
\addplot[black, only marks] table[x=eLP, y=jLP] {results_algorithm1.dat};
\addplot[orange, only marks] table[x=eES2, y=jES2] {results_algorithm1.dat};
\addplot[lime, only marks] table[x=eES2a, y=jES2a] {results_algorithm1.dat};
\legend{PP,ES2,\alg(ES2)}
\end{axis}
\end{tikzpicture}
\caption{Prediction accuracy.}
\label{fig:resultsAlgorithm}
\end{figure}

We investigate the performance of \alg\ algorithm 
with the second order exponential smoothing \alg(ES2), which showed the most promising performance in the previous experiment. We compare the performance of the algorithm \alg(ES2) with applying ES2 and a naive persistent prediction PP stand alone. 
Figure \ref{fig:resultsAlgorithm} plots the errors and jitters on all experiments, one dot represents one dataset and the numerical results are provided in an on-line appendix \footnote{On-line appendix \url{http://datasets.bpti.lt/wp-content/uploads/2013/12/ExSmi-OnlineAppendix.pdf}.}.
We can see from the plot that \alg(ES2) has advantage over simple ES2 in situations where overall error and jitter are quite high, i.e.\ in extremely unpredictable cases. 
This performance supports the intuition, that when an intelligent method cannot do well, it makes sense switching to a robust baseline predictor.
 
Next we analyze this phenomenon in more detail. 
Table \ref{tab:split} provides average errors and jitters for the experiments divided into two groups: (1) difficult to predict identified by high prediction error ($>8$ mm/s) and easy to predict identified by lower prediction error ($\leq 8$ mm/s). 
We see that, indeed, for the difficult to predict cases the algorithm provides a better balance between error and jitter, while it does not disturb much the easier to predict cases.
\begin{table}
\caption{Average performance on difficult and easy to predict cases.}
\label{tab:split}
\centering
\begin{tabularx}{\columnwidth}{Xcccccccc}
\toprule
\multicolumn{1}{c}{\textbf{Group}} &
\textbf{Measure (mm)} &
$\mathit{PP}$ &
$\mathit{ES2}$ &
$\mathit{\alg(ES2)}$\\
\midrule
Difficult 	& average error & 10.8 & 9.6 & 9.1\\ 
$E$					& (std.) & (2.1) & (1.5) & (1.3)\\ 
$>$					& average jitter & 6.0 & 8.8 & 7.0\\
$8$ mm/s					& (std.) & (1.1) & (1.5) & (1.2)\\ 
					& error + jitter & 16.8 & 18.4 & 16.1\\
\midrule
Easy & average error & 7.5 & 4.1 & 4.2\\ 
$E$ 					& (std.) & (3.7) & (1.6) & (1.6)\\ 
$\leq$					& average jitter & 4.1 & 5.0 & 4.8\\
$8$ mm/s					& (std.) & (1.8) & (1.9) & (2.0)\\ 
					& error + jitter & 11.5 & 9.0 & 9.0\\
\bottomrule
\end{tabularx}
\end{table}

Our dataset includes signals with different activities (such as laughing or talking). Next we analyze the performance of \alg\ at different activities. Table \ref{tab:split-activities} presents the results. We can see that normal position demonstrates the lowest overall error and jitter, as it could be expected, since the patient stays still. Prediction in laughing and talking conditions is, hence, more difficult. The proposed \alg\ performs nearly the same as the base model ES2 in normal/other conditions; however, \alg\ consistently performs the best in other than normal conditions, which is a desired feature of our solution. We have implemented outlier control and predictor switch mechanisms so that the predictions stay robust in difficult situations, and these experimental results support that.
\begin{table}
\caption{Average performance with different acitivities.}
\label{tab:split-activities}
\centering
\begin{tabularx}{\columnwidth}{Xcccccccc}
\toprule
\multicolumn{1}{c}{\textbf{Group}} &
\textbf{Measure (mm/s)} &
$\mathit{PP}$ &
$\mathit{ES2}$ &
$\mathit{\alg(ES2)}$\\
\midrule
Normal & average error & 7.0 & 3.9 & 4.0\\ 
					& average jitter & 3.9 & 4.7 & 4.6\\
					& error + jitter & 10.9 & 8.6 & 8.6\\
\midrule
Normal/ & average error & 8.1 & 4.0 & 4.0\\ 
other				& average jitter & 4.2 & 5.0 & 5.0\\
					& error + jitter & 12.3 & 9.0 & 9.0\\
\midrule
Talking & average error & 10.2 & 4.6 & 4.8\\ 
					& average jitter & 5.3 & 6.2 & 6.1\\
					& error + jitter & 15.6 & 10.8 & 10.9\\
\midrule
Talking & average error & 7.6 & 6.1 & 6.1\\ 
					& average jitter & 4.2 & 5.9 & 5.1\\
					& error + jitter & 11.8 & 12.0 & 11.2\\
\midrule
Laughing/& average error & 10.8 & 10.1 & 9.4\\ 
talking 			& average jitter & 6.0 & 9.1 & 7.0\\
					& error + jitter & 16.8 & 19.2 & 16.4\\
\bottomrule
\end{tabularx}
\end{table}

\begin{figure*}
\centering
\includegraphics[width=\textwidth]{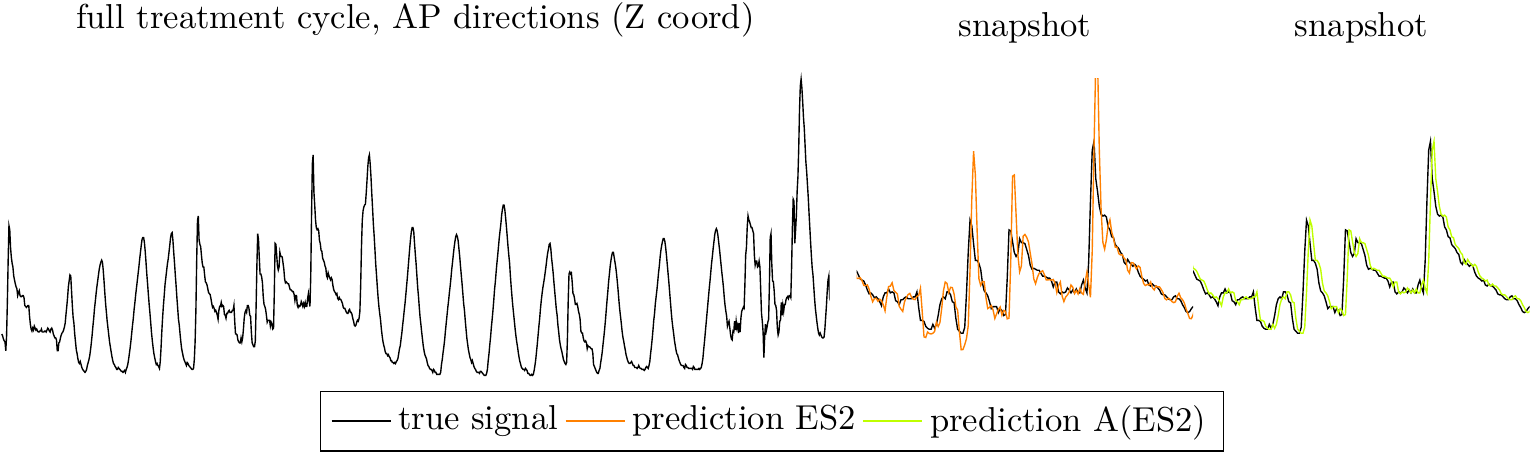}
\caption{Performance on a snapshot of experiment 201205101536-LAC-1-LT-142-6.}
\label{fig:algexample}
\end{figure*}

Finally, Figure~\ref{fig:algexample} plots an example of predictions by the algorithm \alg(ES2) and ES2 stand alone on a difficult to predict case. 
We see that when the true signal suddenly starts to jump ES2 largely overshoots. 
This is because ES2 takes into account the velocity of the signal, observing one sudden jump in the signal level leading to extrapolation of this pattern, 
i.e. predicting that the signal will jump further with a similar speed. 
In such a case when the signal is noisy a naive persistent predictor proves to be more accurate. 
The proposed algorithm combines ES2 and PP and takes advantage of both. 

\section{Discussion}
\label{sec:discussion}

\alg\@, PP and LE approaches have been implemented in a prototype Rubedo system including a HexaPOD couch, 
and an infrared stereocamera (NDI Polaris). This prototype validation has confirmed our experimental results, and several additional observations regarding the performance have been made.
\begin{enumerate}
\item HexaPOD couch is quite sensitive to larger \chblue{speed and direction} \sout{velocity} changes and jitter, i.e. 
the device starts vibrating. Currently, the problem is solved by putting an independent restriction on velocity changes. As a future work, it would be interesting to consider such constraints as part of the prediction algorithm.
\item \alg(ES3) implementation seems to be over-sensitive to periodicity changes, while the period of respiratory motion typically is changing all the time.
That explains why the best results have been achieved by \alg(ES2) 2 and LE with anti-vibration means. 
\end{enumerate}

\chred{In this paper only prediction of tumor motion and its compensation is investigated, but in case of external markers motion of tumor should be predicted from a motion of an external marker, e.g. \cite{KrilaviciusUzupyteZliobaite2013,UzupyteEtAll2015}, which would induce additional error. This technique requires fixing markers near tumor. However they should be out of the beams' way. The couch used in the setup is not constructed to compensate motion, therefore it is not clear how long and how well it would operate over the extended period of time in such a mode.}

\chred{Experiments were performed on motion recorded using external markers under an assumption that tumors move in a similar fashion. Therefore, further investigation with tumor motion could be useful.}

\chred{The important question, which we did not answer in this paper, is how much would a prediction would correct a clinical misalignment of the target? It could be, that linac, MLC, immobilization devices and, especially, a live patient are contribute more the the overall error, while precisions of the most of the existing predictors is sufficient. It is out of scope of this paper, but such analysis could be very interesting.}

In summary, the prototype implementation has demonstrated a promising performance, confirmed our experimental findings, 
and indicated an interesting direction for future investigation.

\section{Conclusions}
\label{sec:Conclusions}

The study investigated prediction models and developed an algorithmic solution \alg\ for predicting the position of a target in 3D in real time. 
\alg\ demonstrated good performance, measured by the prediction accuracy and the jitter of the prediction signal. 
The developed algorithmic solution performs well to be prototyped deployed in radiation therapy applications.

This study has opened several interesting and important directions for future research. 
The first direction is to extend the algorithmic solution \alg\ to take into account technical characteristics of the equipment, for instance, the maximum possible velocity change of the treatment couch. 
While this study treated each respiratory signal as an independent observation, the second interesting and important direction for extension would be to consider multiple signals from different locations simultaneously. 
Taking into account such characteristics it is expected to further improve the precision of treatment.\\

\paragraph{Conflicts of interest}
Tomas Krilavi\v{c}ius, Indr\.e \v{Z}liobait\.e, Henrikas Simonavi\v{c}ius, and  Laimonas Jaru\v{s}evi\v{c}ius declare that they have no conflict of interest.

\paragraph{Acknowledgements} We thank Gabrielius \v{C}aplinskas, Au\v{s}ra Vidugirien\.e and Julius Ruseckas for valuables comments on different aspects of the system. Research was partially funded by ESFA (DADA, VP1-3.1-ŠMM-10-V-02-025).

\section*{References}

\bibliography{bib_repoz}

\appendix

\section{Jitter per time spacing and relative error}

Table \ref{tab:results_rev} presents jitter and error of Figure \ref{fig:results} relative to PP. 
\begin{table}[h]
\caption{Predictive performance of the base models relative to persistent prediction (PP).}
\centering
\begin{tabular}{lcccccc}
\toprule
& PP & LE & MULIN & ES1 & ES2 & ES3 \\
\midrule
relative error & 1.00 & 0.60 & 0.74 & 1.17 & 0.62 & 0.78 \\
relative jitter & 1.00 & 1.45 & 2.06 & 0.92 & 1.27 & 1.22 \\
\bottomrule
\end{tabular}
\label{tab:results_rev}
\end{table}

Table \ref{tab:split_rev} presents jitter and error of Table \ref{tab:split} relative to PP. 
\begin{table}[h]
\caption{Relative performance on difficult to predict and easy to predict cases.}
\label{tab:split_rev}
\centering
\begin{tabularx}{\textwidth}{Xlccc}
\toprule
& Measure & PP & ES2 & \alg(ES2) \\
\midrule
Difficult 	& relative error & 1.00 & 0.89 & 0.84\\ 
$E>8$		& relative jitter & 1.00 & 1.47 & 1.17\\
mm/s			& error + jitter & 1.00 & 1.09 & 0.96\\
\hline
Easy  			& relative error & 1.00 & 0.55 & 0.56\\ 
$E\leq 8$		& relative jitter & 1.00 & 1.21 & 1.18\\
mm/s				& error + jitter & 1.00 & 0.78 & 0.78\\
\bottomrule
\end{tabularx}
\end{table}

\end{document}